\documentclass[letterpaper, 10 pt, conference]{ieeeconf}

\IEEEoverridecommandlockouts
\usepackage{amsmath,amssymb,amsfonts}
\usepackage{amsthm}
\usepackage{array}
\usepackage{algorithm}
\usepackage{algpseudocode}
\usepackage{graphicx}
\usepackage{textcomp}
\usepackage{xcolor}
\usepackage[hidelinks]{hyperref}
\usepackage{accents}
\usepackage{eso-pic}
\usepackage{tikz}
\usetikzlibrary{calc}

\theoremstyle{definition}
\newtheorem{theorem}{Theorem}

\newtheorem{lemma}[theorem]{Lemma}
\newtheorem{proposition}[theorem]{Proposition}

\newtheorem{assumption}[theorem]{Assumption}

\definecolor{mplblue}{HTML}{1F77B4}
\definecolor{mplorange}{HTML}{FF7F0E}
\definecolor{mplgreen}{HTML}{2CA02C}
\definecolor{mplred}{HTML}{D62728}
\definecolor{mplpurple}{HTML}{9467BD}
\definecolor{mplbrown}{HTML}{8C564B}

\newcommand{\legsymwidth}{5mm}
\newcommand{\legsymlw}{1.5pt}
\NewDocumentCommand{\legsym}{ m O{solid} }{%
  \tikz[baseline=-0.6ex]{\draw[#1,#2,line width=\legsymlw] (0,0) -- (\legsymwidth,0);}%
}

\title{\LARGE \bf
Smart Exploration in Reinforcement Learning Using Bounded Uncertainty Models
}

\author{J.S. van Hulst, W.P.M.H. Heemels, D.J. Antunes%
\thanks{The research is carried out as part of the ITEA4 20216 ASIMOV project. The ASIMOV activities are supported by the Netherlands Organisation for Applied Scientific Research TNO and the Dutch Ministry of Economic Affairs and Climate (project number: AI211006). The research leading to these results is partially funded by the German Federal Ministry of Education and Research (BMBF) within the project ASIMOV-D under grant agreement No. 01IS21022G [DLR], based on a decision of the German Bundestag.}%
\thanks{The authors are with the Control Systems Technology Section, Department of Mechanical Engineering, Eindhoven University of Technology, the Netherlands. Emails: {\tt\small\{j.s.v.hulst, m.heemels, d.antunes\}@tue.nl}}%
}

\AddToShipoutPictureBG*{%
	\AtPageUpperLeft{%
		\setlength\unitlength{1in}%
		\hspace*{\dimexpr0.5\paperwidth\relax}%
		\makebox(0,-1)[c]{\begin{tabular}{>{\centering\arraybackslash}p{\textwidth}}
			J.S. van Hulst \emph{et al.}, Smart Exploration in Reinforcement Learning Using Bounded Uncertainty Models, \emph{64th IEEE Conference on Decision and Control (CDC)}, Rio de Janeiro, Brazil, 2025, pp. 5132--5138. DOI: \href{https://doi.org/10.1109/CDC57313.2025.11313018}{10.1109/CDC57313.2025.11313018}. This version contains minor revisions compared to the IEEE publication.
		\end{tabular}}
}}

\AddToShipoutPictureBG*{%
\AtPageUpperLeft{%
\setlength\unitlength{1in}%
\hspace*{\dimexpr0.5\paperwidth\relax}
\makebox(0,-21)[c]{
\footnotesize
\begin{tabular}{>{\centering\arraybackslash}p{\textwidth}}
\textcopyright{} 2025 IEEE. Personal use of this material is permitted. Permission from IEEE must be obtained for all other uses, in any current or future media, including reprinting/republishing this material for advertising or promotional purposes, creating new collective works, for resale or redistribution to servers or lists, or reuse of any copyrighted component of this work in other works.
\end{tabular}}}}

\begin{document}

\maketitle
\thispagestyle{empty}
\pagestyle{empty}

\begin{abstract}
Reinforcement learning (RL) is a powerful framework for decision-making in uncertain environments, but it often requires large amounts of data to learn an optimal policy. We address this challenge by incorporating prior model knowledge to guide exploration and accelerate the learning process. Specifically, we assume access to a model set that contains the true transition kernel and reward function. We optimize over this model set to obtain upper and lower bounds on the Q-function, which are then used to guide the exploration of the agent. We provide theoretical guarantees on the convergence of the Q-function to the optimal Q-function under the proposed class of exploring policies. Furthermore, we also introduce a data-driven regularized version of the model set optimization problem that ensures the convergence of the class of exploring policies to the optimal policy. Lastly, we show that when the model set has a specific structure, namely the bounded-parameter MDP (BMDP) framework, the regularized model set optimization problem becomes convex and simple to implement. In this setting, we also prove finite-time convergence to the optimal policy under mild assumptions. We demonstrate the effectiveness of the proposed exploration strategy, which we call \textbf{BUMEX} (\emph{Bounded Uncertainty Model-based Exploration}), in a simulation study. The results indicate that the proposed method can significantly accelerate learning in benchmark examples. A toolbox is available at \url{https://github.com/JvHulst/BUMEX}.
\end{abstract}

\section{INTRODUCTION}
Reinforcement learning (RL) has shown great success in solving complex decision-making problems in various domains, such as robotics, games, and finance~\cite{Sutton2018}. However, RL often requires large amounts of data to learn an optimal policy, which can be impractical in many real-world applications. This is especially true in environments where data collection is expensive or time-consuming, such as in robotics or healthcare. The need for large datasets comes from the fact that RL algorithms typically explore the environment randomly, causing the agent to visit many suboptimal states and actions~\cite{Kearns2002a}.

A key opportunity arises when one has access to model knowledge, which is often the case in many scenarios. This information is typically not utilized in traditional RL algorithms. For instance, we may know that certain states cannot be reached from particular other states or that there is a specific goal state. In this paper, we propose using such available knowledge, giving information on the set of models to which the true model belongs, and using this to guide the agent's exploration, thus speeding up the learning process.

More specifically, we propose estimating bounds on the Q-function using optimistic/pessimistic optimization over the available model set. We then use these bounds to guide the agent's exploration by prioritizing uncertain or promising regions of the state/action space. Additionally, we propose regularizing the model set optimization problem using a database of observed transitions. We show that the Q-function converges to the optimal Q-function under the exploring policy and that the exploring policy converges to the optimal policy in the regularized case. Moreover, for a specific model set characterization we obtain a practical algorithm with a convex (regularized) model set optimization. We refer to this practical algorithm as BUMEX (\emph{Bounded Uncertainty Model-based Exploration}).

This work is at the intersection of model-based RL and smart exploration in RL. We will now briefly review the relevant literature, highlighting the relation and differences to the current work.

Model-based reinforcement learning is a popular approach to speed up the learning process in RL by incorporating model knowledge. In traditional model-based RL, the agent learns a model of the environment and uses this model to plan its actions~\cite{Sutton1991,Deisenroth2011}. In our work, we use prior model knowledge to guide the exploration of the agent, rather than to plan its actions.
Smart exploration in RL focuses on how to explore the environment efficiently to speed up learning. Several prominent approaches use confidence bounds on value functions to guide exploration, including Upper Confidence Bound methods~\cite{Kearns2002a,Jin2018}, UCRL2~\cite{Jaksch2010}, and Posterior Sampling for Reinforcement Learning (PSRL)~\cite{Osband2013}. Bayesian RL builds on this by using Bayesian inference to guide the agent's exploration~\cite{Ghavamzadeh2016}. These methods typically rely on uncertainty estimates derived from a single nominal model or confidence intervals around model parameters.
In contrast, our work employs a set of models to direct exploration, rather than relying on a nominal model with uncertainty estimates.

Our work builds most directly on the bounded-parameter MDP (BMDP) framework~\cite{Givan2000}, which characterizes model uncertainty through intervals on transition probabilities and rewards. BMDPs are closely related to uncertain MDPs (uMDPs) and interval MDPs (IMDPs). The BMDP framework has been used to develop algorithms that are robust to model uncertainty~\cite{Haddad2018}. The present work builds upon this idea by using the BMDP framework to guide the exploration of the agent. Additionally, our exploration strategy draws inspiration from Bayesian optimization~\cite{Garnett2023}, which balances exploration and exploitation by using acquisition functions. We adapt similar heuristics for selecting actions based on Q-function bounds.

The main contributions can be summarized as follows.
\begin{itemize}
    \item We introduce an exploration strategy for reinforcement learning that leverages bounds on the Q-function obtained from a model set. We provide theoretical guarantees on the convergence of the Q-function to the optimal Q-function under the proposed exploration strategy.
    \item We propose a data-regularized version of the model set optimization problem that ensures convergence of the Q-function bounds to the optimal Q-function.
    \item We propose a practical algorithm, BUMEX, for the BMDP framework and establish, under specific conditions, finite-time convergence to the optimal policy.
\end{itemize}
Our main theoretical results focus on finite state-action spaces, which allows us to provide (finite-time) convergence guarantees and construct a practical algorithm.

The paper is organized as follows. Section~\ref{sec:problem} introduces the problem formulation.  Section~\ref{sec:method} presents the general framework for the proposed method and the technical results. Section~\ref{sec:practical_algorithm} discusses a practical algorithm that leverages the theoretical results. Section~\ref{sec:results} presents simulations and Section~\ref{sec:conclusions} concludes the paper.

\section{PROBLEM FORMULATION}\label{sec:problem}
We model the decision-making environment as a Markov decision process (MDP), which we define in a general setting to encompass both finite and infinite (continuous) state and action spaces. We assume an infinite-horizon discounted MDP, where the agent interacts with the environment over an infinite number of time steps. This MDP is defined as the tuple $\mathcal{M} = (\mathcal{X}, \mathcal{U}, P, g, \gamma)$, with the following ingredients:
\begin{itemize}
    \item \textbf{State Space} $\mathcal{X}$ is a measurable (Polish) space representing the set of possible states of the environment. We denote the Borel $\sigma$-algebra on $\mathcal{X}$ by $\mathcal{B}(\mathcal{X})$.
    \item \textbf{Action Space} $\mathcal{U}$ is a measurable (Polish) space representing the set of inputs or actions that the agent can take. We denote the Borel $\sigma$-algebra on $\mathcal{U}$ by $\mathcal{B}(\mathcal{U})$.
    \item \textbf{Transition Kernel} $P: \mathcal{X} \times \mathcal{U} \times \mathcal{B}(\mathcal{X}) \to [0,1]$ is a Markovian transition kernel, where $P(x,u,\cdot)$ is a probability measure on $\mathcal{X}$ for all $(x,u) \in \mathcal{X} \times \mathcal{U}$. We assume that for every bounded measurable function $h:\mathcal{X}\to\mathbb{R}$, the mapping $(x,u) \mapsto \int_{\mathcal{X}} h(x') P(x,u,dx')$ is measurable.
    \item \textbf{Reward Function} $g: \mathcal{X} \times \mathcal{U} \to \mathbb{R}$ is a bounded, measurable function, where $g(x,u)$ is the reward obtained by the agent when taking action $u$ in state $x$. We assume deterministic rewards for simplicity. The extension to stochastic rewards is straightforward.
    \item \textbf{Discount Factor} $\gamma \in [0,1)$ is the discount factor that determines the importance of future rewards.
\end{itemize}
The goal in this setting is to find a policy $\pi \in \Pi$, where $\Pi := \{ \pi: \mathcal{X} \to \text{Prob}(\mathcal{U}) \mid \pi \text{ is measurable} \}$, such that the expected cumulative reward or value is maximized. Here $\text{Prob}(\mathcal{U})$ denotes the set of Borel probability measures on $(\mathcal{U}, \mathcal{B}(\mathcal{U}))$. The value function $V^\pi: \mathcal{X} \to \mathbb{R}$ of a policy $\pi \in \Pi$ is defined as
\begin{equation}
    \label{eq:policy_value_function}
    V^\pi(x) = \mathbb{E}^\pi \left[ \sum_{k=0}^\infty \gamma^k g(x_k,u_k) \,\bigg|\, x_0 = x \right],
\end{equation}
where $(x_k,u_k)$ denotes the state-action pair at time $k$ under policy $\pi$, and $\mathbb{E}^\pi$ denotes the expectation with respect to the probability measure induced by $\pi$ and the transition kernel $P$. A policy's Q-function $Q^\pi: \mathcal{X} \times \mathcal{U} \to \mathbb{R}$ is defined as
\begin{equation}
    \label{eq:policy_Q_function}
    Q^\pi(x,u) = g(x,u) + \gamma \int_{\mathcal{X}} V^\pi(x') P(x,u,dx'),
\end{equation}
where $Q^\pi(x,u)$ is the expected cumulative reward of taking action $u$ in state $x$, then following policy $\pi$. We assume that for every $\pi \in \Pi$, the integrals in~\eqref{eq:policy_value_function},~\eqref{eq:policy_Q_function} are well-defined. The optimal Q-function $Q^*: \mathcal{X} \times \mathcal{U} \to \mathbb{R}$ is defined as
\begin{equation}
    \label{eq:optimal_Q_function}
    Q^*(x,u) = \sup_{\pi \in \Pi} Q^\pi(x,u).
\end{equation}
An optimal policy $\pi^*$ can be obtained from the optimal Q-function as
$\pi^*(x) \in \arg\max_{u \in \mathcal{U}} Q^*(x,u)$, whenever this maximum is attained. The optimal Q-function can also be defined in terms of the Bellman operator $\mathcal{T}$ as $Q^* = \mathcal{T} Q^*$, where $\mathcal{T}: \mathcal{B}(\mathcal{X} \times \mathcal{U}) \to \mathcal{B}(\mathcal{X} \times \mathcal{U})$ is defined as
\begin{equation*}
    \label{eq:Bellman_operator}
    (\mathcal{T}Q)(x,u) = g(x,u) + \gamma \int_{\mathcal{X}} \sup_{u' \in \mathcal{U}} Q(x',u') P(x,u,dx').
\end{equation*}

As usual in reinforcement learning, our objective is to find the optimal policy $\pi^*$ by interacting with the environment and learning from the observed data $\mathcal{D}:= {\{} (x_k,u_k,x_{k+1}) {\}}_{k=0}^{N-1}$. The transition kernel $P$ and the reward function $g$ are assumed to be unknown to the agent. However, we assume that the agent has access to a model set $\hat{\mathcal{M}}$ that contains the true transition kernel and reward function. This is formalized in the following assumption.
\begin{assumption}
    The model set $\hat{\mathcal{M}}$ is defined as $\hat{\mathcal{M}} = (\mathcal{X}, \mathcal{U}, \mathcal{P}, \mathcal{G}, \gamma)$, where $\mathcal{P}$ is a set of transition kernels and $\mathcal{G}$ is a set of reward functions. The true transition kernel $P$ belongs to $\mathcal{P}$, and the true reward function $g$ belongs to $\mathcal{G}$. The model set $\hat{\mathcal{M}}$ is known to the agent.
\end{assumption}
Thus, the problem is to find the optimal policy $\pi^*$ based on the model set $\hat{\mathcal{M}}$ and the observed data $\mathcal{D}$.

\section{PROPOSED METHOD}\label{sec:method}
We will first discuss a version of the proposed method that fits the general MDP setting introduced in the previous section. This version cannot be directly applied in practice because optimization over the general model set is intractable. This version presents the theoretical foundation for a more practical version that we will detail in Section~\ref{sec:practical_algorithm}.

\subsection{General Framework}
Given the model set, we can determine bounds on the Q-function that hold for all models in the set. These bounds are constructed using the Bellman updates, given by
\begin{equation}
    \label{eq:general_Q-bound_updates}
    \begin{aligned}
        (\underaccent{\bar}{\mathcal{T}}\underaccent{\bar}{Q})(x,u) &= \inf_{g \in \mathcal{G}} g(x,u) + \gamma \inf_{P \in \mathcal{P}} \int_{\mathcal{X}} \underaccent{\bar}{V}(x') P(x,u,dx'), \\
        (\bar{\mathcal{T}}\bar{Q})(x,u) &= \sup_{g \in \mathcal{G}} g(x,u) + \gamma \sup_{P \in \mathcal{P}} \int_{\mathcal{X}} \bar{V}(x') P(x,u,dx'),
    \end{aligned}
\end{equation}
in which $\underaccent{\bar}{V}$ and $\bar{V}$ are defined as
\begin{equation}
    \label{eq:V_bounds}
        \underaccent{\bar}{V}(x) := \sup_{u \in \mathcal{U}} \underaccent{\bar}{Q}(x,u), \quad \bar{V}(x) := \sup_{u \in \mathcal{U}} \bar{Q}(x,u).
\end{equation}
These relationships and the resulting bounds are related to adversarial MDPs, where the agent aims to minimize the reward in the worst-case scenario, see e.g.~\cite{Russel2019}. We assume throughout that the functions of $(x,u)$ defined by the infima and suprema in~\eqref{eq:general_Q-bound_updates} are measurable, so that $\underaccent{\bar}{\mathcal{T}}$ and $\bar{\mathcal{T}}$ map $\mathcal{B}(\mathcal{X} \times \mathcal{U})$ into itself. This holds automatically when $\mathcal{X}$ and $\mathcal{U}$ are finite, which is the setting of every convergence result in this paper. In the general case, sufficient conditions follow from the measurable selection theory of~\cite[Ch.~7]{bertsekas1996a}.

\begin{lemma}\label{lem:Q_function_bounds}
    Let $\underaccent{\bar}{\mathcal{T}}$ and $\bar{\mathcal{T}}$ denote the Bellman operators defined in~\eqref{eq:general_Q-bound_updates}. These operators are contraction mappings on $\mathcal{B}(\mathcal{X} \times \mathcal{U})$ with unique fixed points $\underaccent{\bar}{Q}^*$ and $\bar{Q}^*$, respectively. Then
    \begin{equation}
        \label{eq:Q_function_bounds}
        \underaccent{\bar}{Q}^*(x,u) \leq Q^*(x,u) \leq \bar{Q}^*(x,u), \quad \forall (x,u) \in \mathcal{X} \times \mathcal{U},
    \end{equation}
    where $Q^*$ is the optimal Q-function defined in~\eqref{eq:optimal_Q_function}. Similarly, the corresponding value functions
    \begin{equation*}
        \label{eq:optimal_V_bounds}
        \underaccent{\bar}{V}^*(x) := \sup_{u \in \mathcal{U}} \underaccent{\bar}{Q}^*(x,u) \quad \text{and} \quad \bar{V}^*(x) := \sup_{u \in \mathcal{U}} \bar{Q}^*(x,u),
    \end{equation*}
    satisfy
    $\underaccent{\bar}{V}^*(x) \leq V^*(x) \leq \bar{V}^*(x), \quad \forall x \in \mathcal{X}$,
    where $V^*(x):=\sup_{u \in \mathcal{U}} Q^*(x,u)$ is the optimal value function.
\end{lemma}
\begin{proof}
    Since the reward function $g$ is bounded and $\gamma\in[0,1)$, the operators $\underaccent{\bar}{\mathcal{T}}$ and $\bar{\mathcal{T}}$ are contraction mappings with modulus $\gamma$ with respect to the supremum norm. By the Banach fixed-point theorem, each operator has a unique fixed point. Since the true MDP $\mathcal{M}$ is contained in the model set $\hat{\mathcal{M}}$, we have $\underaccent{\bar}{\mathcal{T}} Q \leq \mathcal{T} Q \leq \bar{\mathcal{T}} Q$ for any Q-function $Q$. The monotonicity of these operators implies the ordering of their fixed points as in~\eqref{eq:Q_function_bounds}, and the value function bounds follow directly.
\end{proof}
The paper~\cite{Givan2000} on bounded-parameter MDPs establishes similar bounds for the value function in finite state spaces using interval-based model sets. Our result generalizes this by considering both finite and infinite spaces and by using Q-function bounds instead. Note that the initialization of the Q-bounds does not affect the sandwiching property that is obtained upon convergence. Typically, we initialize the Q-bounds using prior knowledge or heuristics about the environment to speed up the convergence.

As stated next, the guaranteed Q-bounds can be used to test if certain inputs are suboptimal or optimal.
\begin{proposition}\label{prop:suboptimality_inputs}
    Let $\underaccent{\bar}{Q}^*$ and $\bar{Q}^*$ be guaranteed lower and upper bounds on the optimal Q-function $Q^*$, respectively, i.e.,~\eqref{eq:Q_function_bounds} holds. Then, the input $u$ at state $x$ is suboptimal if
    \begin{equation}
        \label{eq:suboptimality_input}
        \bar{Q}^*(x,u) < \underaccent{\bar}{V}^*(x),
    \end{equation}
    and the input $u$ at state $x$ is optimal if
    \begin{equation}
        \label{eq:optimality_input}
        \underaccent{\bar}{Q}^*(x,u) \geq \sup_{v \in \mathcal{U} \setminus \{u\}} \bar{Q}^*(x,v).
    \end{equation}
\end{proposition}
This result follows directly from the guaranteed Q-bounds.

In the next section, we will discuss an exploration strategy that leverages the Q-function bounds to guide the agent's exploration in the environment.

\subsection{Exploration Strategy}\label{sec:exploration}
We propose a class of exploring policies to be used during the acquisition of the data that leverages the Q-function bounds. The exploring policy randomly selects the input $u$ at state $x$ according to weighted random sampling with input-dependent weights
{\begingroup\fontsize{8.9pt}{10}\selectfont
\begin{equation}\label{eq:exploring_policy_weights}
	\phi(x,u) =  \begin{cases}
        \xi, & \text{if } \underaccent{\bar}{Q}(x,u) \geq \sup_{v\in\mathcal{U}\setminus \{u\}}\bar{Q}(x,v),\\
        \beta(x,u), & \text{if } \underaccent{\bar}{Q}(x,u)\neq\bar{Q}(x,u) \text{ and } \bar{Q}(x,u) > \underaccent{\bar}{V}(x),\\
        0, & \text{if } \underaccent{\bar}{Q}(x,u) = \bar{Q}(x,u) \text{ and } \bar{Q}(x,u) \leq \underaccent{\bar}{V}(x), \\
        \zeta, & \text{if } \underaccent{\bar}{Q}(x,u) \neq \bar{Q}(x,u) \text{ and } \bar{Q}(x,u) \leq \underaccent{\bar}{V}(x), \\
        \zeta, & \text{if } \underaccent{\bar}{Q}(x,u) = \bar{Q}(x,u) \text{ and } \bar{Q}(x,u) > \underaccent{\bar}{V}(x),
	\end{cases}
\end{equation}
\endgroup}
with $\phi: \mathcal{X} \times \mathcal{U} \to \mathbb{R}_{\geq 0}$ the sampling weight of input $u$ at state $x$. Here, $\xi \in \mathbb{R}_{> 0}$ and $\zeta \in \mathbb{R}_{\geq 0}$ are constants, and $\beta: \mathcal{X} \times \mathcal{U} \to \mathbb{R}_{> 0}$. The conditions are evaluated in the order listed, and the first one that applies determines the weight. The induced exploration policy $\pi^\phi$ is defined by
\begin{equation}
    \label{eq:exploring_policy}
    \pi^\phi(x,u) = \frac{\phi(x,u)}{\int_{\mathcal{U}} \phi(x,v)\,\mu(dv)},
\end{equation}
with $\mu$ a suitable reference measure on $\mathcal{U}$ (e.g., the counting measure if $\mathcal{U}$ is finite).

The first case in~\eqref{eq:exploring_policy_weights} implies that the input $u$ at state $x$ is guaranteed to be optimal (by {Proposition~\ref{prop:suboptimality_inputs}}). For this reason, we assign it an arbitrary positive weight $\xi$. Note that if an input satisfies this condition, this rules out the possibility that any other inputs satisfy the second condition in~\eqref{eq:exploring_policy_weights}.

The second case implies that the quality of input $u$ at state $x$ is uncertain but that the input has the potential to improve $\underaccent{\bar}{V}$. We assign this input a positive weight $\beta$ that depends on $x$ and $u$. This allows us to potentially assign more weight to the input if it is highly uncertain or has a high Q-value upper bound $\bar{Q}$. Section~\ref{subsec:BO_weighting} provides heuristic guidance on the choice of $\beta$, inspired by Bayesian optimization.

\begin{figure}[!t]
    \centering
    \vspace{0.25cm}
    \includegraphics[width=80mm]{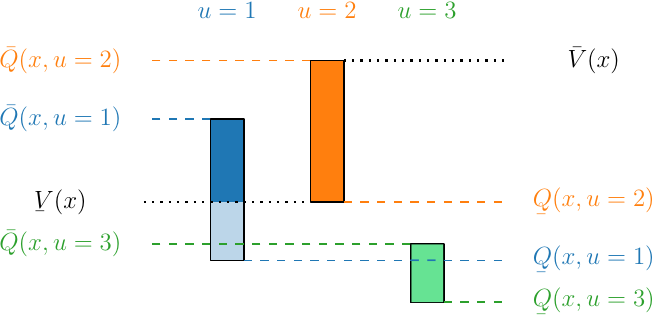}
    \caption{Schematic representation of Q-function bounds. From this particular example, using {Proposition~\ref{prop:suboptimality_inputs}} we can deduce that the choice $u=3$ is guaranteed to be suboptimal for this state $x$, as its upper bound Q-value $\bar{Q}(x,3)$ is lower than $\protect\underaccent{\bar}{V}(x)$. By~\eqref{eq:exploring_policy_weights}, the input choice $u=3$ is therefore assigned a weight of $\zeta$. In contrast, actions $u=1$ and $u=2$, for which the bounds do not decisively rule out optimality, receive weight $\beta(x,u)$. Since $\protect\underaccent{\bar}{Q}(x,1)$ is below $\protect\underaccent{\bar}{V}(x)$, the proposed heuristics in Section~\ref{subsec:BO_weighting} explore the input $u=2$ more frequently.}\label{fig:exploring_policy}
\end{figure}

The third case in~\eqref{eq:exploring_policy_weights} assigns zero weight to actions with tight Q-bounds that are guaranteed to be suboptimal. The fourth case assigns a nonnegative weight of $\zeta$ to inputs which are uncertain yet appear to be suboptimal, while the fifth case assigns the same weight to inputs which are certain and not guaranteed to be suboptimal. Note that in many cases, we can select $\zeta=0$ to prune such inputs. However, this pruning might be too aggressive in some cases if we introduce data-driven regularization of the model set optimization, which will be discussed later in this paper. Fig.~\ref{fig:exploring_policy} illustrates the exploring policy based on the bounds.

As shown next, the convergence properties of standard Q-learning are retained by the proposed exploring policy, as the exploration is based on guaranteed bounds.
\begin{theorem}\label{thm:exploration_convergence}
    Consider finite state-action spaces. Let $Q$ be the Q-function obtained using standard Q-learning
    {\small
    \begin{equation}\label{eq:Q-learning}
        Q(x_k,u_k) \leftarrow (1-\alpha_k) Q(x_k,u_k) + \alpha_k \left[ r_k + \gamma \max_{u' \in \mathcal{U}} Q(x_{k+1},u') \right],
    \end{equation}
    }
    under the exploring policy $\pi^\phi$ defined in~\eqref{eq:exploring_policy} with parameters $\xi \in \mathbb{R}_{>0}, \zeta \in \mathbb{R}_{\geq 0}$ and function $\beta: \mathcal{X} \times \mathcal{U} \to \mathbb{R}_{>0}$, initialized such that $\underaccent{\bar}{Q} \leq Q \leq \bar{Q}$. If every state is visited infinitely often (through e.g., exploring starts), and the learning rate $\alpha_k$ satisfies the Robbins-Monro conditions $\sum_{k=0}^\infty \alpha_k = \infty$ and $\sum_{k=0}^\infty \alpha_k^2 < \infty$, then the greedy policy with respect to $Q$ converges to the optimal policy with probability 1 as $k \to \infty$.
\end{theorem}
\begin{proof}
    Standard Q-learning convergence requires that every state-action pair is updated infinitely often. Our exploring policy $\pi^\phi$ prunes actions that are guaranteed to be suboptimal by~\eqref{eq:suboptimality_input}, meaning not every state-action pair is updated. However, the key insight is that for every state $x$, every action that could potentially be optimal is still updated infinitely often (via exploring starts). Thus, while $Q(x,u)$ may not converge to $Q^*(x,u)$ for pruned actions, the greedy policy $\pi(x) \in \arg\max_{u \in \mathcal{U}} Q(x,u)$ converges to the optimal policy $\pi^*(x)$ with probability 1, since the Q-values of all candidate optimal actions either converge to their true values or are dominated by actions that do.
\end{proof}

In the next section, we detail how to select the weighting function $\beta$ using several heuristics.

\subsubsection{Choice of Weighting $\beta$}\label{subsec:BO_weighting}
Inspired by acquisition functions in Bayesian optimization~\cite{Garnett2023}, we can select efficient weighting functions $\beta$. We are interested in improving the best worst-case Q-value $\underaccent{\bar}{V}(x)$ at state $x$. This improvement as a function of the choice of input $u$ is given by
\begin{equation}\label{eq:improvement}
    I(x,u) = \max \left(0, q-\underaccent{\bar}{V}(x)\right),
\end{equation}
in which $q$ denotes the unknown Q-value of input $u$ at state $x$, which lies between $\underaccent{\bar}{Q}(x,u)$ and $\bar{Q}(x,u)$.
By assuming that the probability of Q-values within the bounds $\underaccent{\bar}{Q}$ and $\bar{Q}$ is uniform, we can calculate the probability that a given input $u$ has an improvement greater than 0. This \emph{probability of improvement} is given by
\begin{equation}
\label{eq:PI}
    \beta(x,u) = \text{Prob}[I(x,u)> 0]
    = \frac{ \max \left(0, \bar{Q}(x,u)-\underaccent{\bar}{V}(x)\right) }{ \bar{Q}(x,u)-\underaccent{\bar}{Q}(x,u) }.
\end{equation}
Alternatively, we can take the \emph{expected} value of the \emph{improvement}~\eqref{eq:improvement} to obtain
\begin{equation}
\label{eq:EI}
\begin{aligned}
    \beta(x,u) = \mathbb{E}[I(x,u)]
    = \frac{ \left(\max \left(0, \bar{Q}(x,u)-\underaccent{\bar}{V}(x)\right)\right)^2 }{ 2\left(\bar{Q}(x,u)-\underaccent{\bar}{Q}(x,u) \right)}.
\end{aligned}
\end{equation}
These sampling weights prioritize exploring inputs with a large region of their Q-value uncertainty range higher than the current best-worst-case input. Note that we can assume other probability distributions (with bounded support) for the Q-values between $\underaccent{\bar}{Q}$ and $\bar{Q}$. One such example that generalizes the uniform distribution is the Bates distribution, which allows us to assume a higher relative probability concentration around the midpoint of the Q-value bounds.

In the next section, we detail an approach to include data in the method by regularizing the optimization over the model set $\mathcal{P}$ to bias it toward the observed transitions.

\subsection{Regularization with Observed Data}\label{sec:regularization}
We propose that by regularizing the optimization over the model set using observed data $\mathcal{D}$, the bounds converge to the optimal Q-function under mild conditions. We can re-define the Q-bound Bellman updates~\eqref{eq:general_Q-bound_updates} as
\begin{equation}
    \label{eq:regularized_Q-bound_updates}
    \begin{aligned}
        (\underaccent{\bar}{\mathcal{T}}\underaccent{\bar}{Q})(x,u) &= \inf_{g \in \mathcal{G}} g(x,u) + \gamma \int_{\mathcal{X}} \underaccent{\bar}{V}(x') \underaccent{\bar}{P}^*(x,u,dx'), \\
        (\bar{\mathcal{T}}\bar{Q})(x,u) &= \sup_{g \in \mathcal{G}} g(x,u) + \gamma \int_{\mathcal{X}} \bar{V}(x') \bar{P}^*(x,u,dx'),
    \end{aligned}
\end{equation}
with
{\begingroup\fontsize{8.5pt}{10}\selectfont
\begin{equation}
    \label{eq:regularized_model_set_optimization}
    \begin{aligned}
        \underaccent{\bar}{P}^*(x,u,dx') &:= \arg\min_{P \in \mathcal{P}} \int_{\mathcal{X}} \underaccent{\bar}{V}(x') P(x,u,dx') + \lambda(x,u) d(P,P_E), \\
        \bar{P}^*(x,u,dx') &:= \arg\max_{P \in \mathcal{P}} \int_{\mathcal{X}} \bar{V}(x') P(x,u,dx') - \lambda(x,u) d(P,P_E),
    \end{aligned}
\end{equation}
\endgroup}
where we assume that the minimum and maximum exist (which is the case if for instance $\mathcal{P}$ is compact). Here, $P_E: \mathcal{X} \times \mathcal{U} \times \mathcal{B}(\mathcal{X}) \to [0,1]$ denotes the empirical transition kernel obtained from the observed data $\mathcal{D}$, and $\lambda: \mathcal{X} \times \mathcal{U} \to \mathbb{R}_{\geq 0}$ is a regularization parameter. The function $d(P,P_E)$ is a divergence between the transition kernel $P$ and the empirical transition kernel $P_E$. The divergence is used to bias the optimized transition kernel toward the observed data, while still being within the model set. The function $d$ must satisfy $d(P, P_E) \geq 0$, with equality if and only if $P = P_E$.
The variable $\lambda$ determines the trade-off between the model set and the observed data. It should increase with the size of $\mathcal{D}$ to ensure that the selected model from the set is biased toward the observed data, such that
\begin{equation}
    \label{eq:lambda_limit}
    \lim_{T(x,u) \to \infty} \lambda(x,u) = \infty,
\end{equation}
where $T(x,u,x')$ is the number of observed transitions from state $x$ under action $u$ to state $x'$ in the dataset $\mathcal{D}$, and $T(x,u) := \sum_{x'} T(x,u,x')$. In practice, $\lambda$ can be chosen as any increasing function of the visit count $T(x,u)$ that satisfies~\eqref{eq:lambda_limit}.

Since the rewards are deterministic, whenever a reward $r_k$ is observed at state-action pair $(x_k,u_k)$, the set of reward functions $\mathcal{G}$ can be constrained to include only elements that correctly assign this reward at that state-action pair, i.e.,
\begin{equation}
    \label{eq:reward_set_shrinking}
    \mathcal{G} = \{g \in \mathcal{G} \mid g(x_k,u_k) = r_k\}.
\end{equation}

Together with the regularized model set optimization, we obtain a convergence result which is formalized next.
\begin{proposition}\label{prop:regularized_convergence}
    Assuming finite state-action spaces, let $\lambda: \mathcal{X} \times \mathcal{U} \to \mathbb{R}_{\geq 0}$ be a regularization parameter satisfying~\eqref{eq:lambda_limit}, and let $d(P,P_E)$ be a divergence such that $d(P,P_E)\geq0$ and $d(P,P_E)=0$ if and only if $P=P_E$. Let $\mathcal{P}$ and $\mathcal{G}$ be the set of transition kernels and the set of reward functions, respectively. Assume that the reward function is deterministic. If every state-action pair is visited infinitely often, then,
    \begin{equation}
        \label{eq:regularized_convergence}
        \lim_{N \to \infty} \underaccent{\bar}{Q} = \lim_{N \to \infty} \bar{Q} = Q^*.
    \end{equation}
\end{proposition}
\begin{proof}
    Since every state-action pair is visited infinitely often, $\lambda(x,u)\to\infty$ by~\eqref{eq:lambda_limit}. The regularization term then forces the optimization to select $P=P_E$, as the regularization dominates the objective. With deterministic rewards, $\mathcal{G}$ reduces to the true reward function after one visit per state-action pair through~\eqref{eq:reward_set_shrinking}. By the law of large numbers, $P_E$ converges to the true transition kernel $P$. As the Bellman operator is a contraction (see~\cite{Bertsekas2022}), it follows that both the lower and upper Q-function bounds converge to $Q^*$.
\end{proof}

Note that for any value of $\lambda$, the Q-bound updates in~\eqref{eq:general_Q-bound_updates} are still convergent. However, it should be noted that the resulting converged Q-bounds are only guaranteed to bound the optimal Q-function $Q^*$ if $\lambda=0$ (by {Lemma~\ref{lem:Q_function_bounds}}) or if $P_E=P$ and $\lambda=\infty$ (by {Proposition~\ref{prop:regularized_convergence}}).

\begin{algorithm}[!b]
    \caption{\textbf{BUMEX:} Q-learning with exploring policy and regularized model set optimization}
    \label{alg:Q-learning_regularized}
    \begin{algorithmic}
        \State Calculate Q-function bounds $\underaccent{\bar}{Q}$ and $\bar{Q}$ using~\eqref{eq:general_Q-bound_updates}.
        \State Initialize $Q$ such that $\underaccent{\bar}{Q} \leq Q \leq \bar{Q}$.
        \For{Episode $e = 1,2,3,\dots$}
            \State Initialize state $x_0$.
            \For{$k = 0,1,2,\dots$ until terminal}
                \If{$\text{rand}() < \epsilon$}
                    \State Select $u_k$ according to exploring policy $\pi^\phi$.
                \Else
                    \State Select $u_k = \arg\max_{u \in \mathcal{U}} Q(x_k,u)$.
                \EndIf
                \State Apply $u_k$ and observe $r_k$ and $x_{k+1}$.
                \State Reduce the reward function set $\mathcal{G}$ such that $g(x_k,u_k) = r_k$ for all $g \in \mathcal{G}$.
                \State Update $Q$ using $(x_k, u_k, r_k, x_{k+1})$ and~\eqref{eq:Q-learning}.
            \EndFor
            \If{$e \mod L = 0$}
                \State Update the Q-bounds using~\eqref{eq:regularized_Q-bound_updates} with~\eqref{eq:regularized_model_set_optimization}.
                \State Clip the Q-function $Q$ to the bounds $\underaccent{\bar}{Q}, \bar{Q}$.
            \EndIf
        \EndFor
    \end{algorithmic}
\end{algorithm}

The regularized model set optimization is incorporated into our proposed method, which is summarized in Algorithm~\ref{alg:Q-learning_regularized}, where the data-regularized Q-bound iterations are performed every $L$ episodes. To improve learning efficiency, we employ an $\epsilon$-greedy type strategy, balancing exploration/exploitation. Additionally, we clip the Q-function to the bounds whenever they are updated. By combining our previous results, we obtain the following convergence guarantee for the exploring policy.

\begin{theorem}\label{thm:exploration_and_regularization_convergence}
    Assume finite state and action spaces. Let $\pi^\phi$ be the exploring policy defined in~\eqref{eq:exploring_policy} for which we additionally assume that $\zeta$ is positive. Let $\lambda$ satisfy~\eqref{eq:lambda_limit}. Assume that $d(P,P_E)$ is a divergence with $d(P,P_E)\geq 0$ and $d(P,P_E)=0$ if and only if $P=P_E$. Assume deterministic rewards, and that every state is visited infinitely often. Then, under {Algorithm~\ref{alg:Q-learning_regularized}}, the Q-bounds converge to the optimal Q-function and the exploring policy $\pi^\phi$ converges to the optimal policy with probability 1 as $k\to\infty$.
\end{theorem}
\begin{proof}
    With $\zeta > 0$, the exploring policy assigns positive weight to all actions except those that are certain and guaranteed suboptimal. For uncertain state-action pairs, the policy ensures infinite exploration, causing $\lambda(x,u)\to\infty$ and forcing convergence to the empirical kernel. By Proposition~\ref{prop:regularized_convergence}, the Q-bounds converge to $Q^*$. The exploring policy then selects only guaranteed optimal actions, converging to the optimal policy.
\end{proof}

\section{PRACTICAL ALGORITHM: BUMEX}\label{sec:practical_algorithm}
In this section, we consider the finite state and action space case, where the MDP transitions are reduced to a transition probability tensor $M$, such that
\begin{equation}
    \text{Prob}(x_{k+1} = x' \mid x_k = x, u_k = u) = M(x,u,x'),
\end{equation}
with $M \in [0,1]^{|\mathcal{X}| \times |\mathcal{U}| \times |\mathcal{X}|}$ and $\sum_{x' \in \mathcal{X}} M(x,u,x') = 1$ for all $(x,u) \in \mathcal{X} \times \mathcal{U}$. The MDP model set $\hat{\mathcal{M}}$ is defined as a bounded-parameter MDP model~\cite{Givan2000}. The set that the true transition kernel $M$ belongs to is defined as
{\small
\begin{equation}
    \label{eq:BMDP_probability_transitions}
        \mathcal{P} = \bigg\{M ~\big|~ \underaccent{\bar}{M} \leq M \leq \bar{M}, \sum_{x'\in \mathcal{X}} M(x,u,x') = 1, \quad \forall~ x,u \bigg\},
\end{equation}}
for known, well-defined lower and upper bounds $\underaccent{\bar}{M}, \bar{M} \in [0,1]^{|\mathcal{X}| \times |\mathcal{U}| \times |\mathcal{X}|}$.
The reward function set that the true reward function $g$ belongs to is defined as
$\mathcal{G} = \left\{g ~\big|~ \underaccent{\bar}{g} \leq g \leq \bar{g}, \quad \forall~ x,u \right\}$.
Thus, we assume that the true transition kernel and reward function are bounded by known upper and lower bounds.

The function $d$ is chosen to be the Kullback-Leibler (KL) divergence between the transition kernel and the observed data, given by
\begin{equation}
    \label{eq:KL_divergence}
    d(M,M_E) = \sum_{x' \in \mathcal{X}} M(x,u,x') \log \left( \frac{M(x,u,x')}{M_E(x,u,x')} \right),
\end{equation}
where $M_E$ is the empirical transition probability tensor from the observed data $\mathcal{D}$ through $M_E(x,u,x') = \frac{T(x,u,x')}{T(x,u)}$. We choose the KL-divergence since it satisfies $d(P,P_E)\geq0$ with equality if and only if $P=P_E$, is convex, and is a natural choice for comparing probability distributions.

The resulting Bellman updates are given by
\begin{equation}
    \begin{aligned}
        (\underaccent{\bar}{\mathcal{T}}\underaccent{\bar}{Q})(x,u) &= \underaccent{\bar}{g}(x,u) + \gamma \sum_{x' \in \mathcal{X}} \underaccent{\bar}{V}(x') \underaccent{\bar}{M}^*(x,u,x'), \\
        (\bar{\mathcal{T}}\bar{Q})(x,u) &= \bar{g}(x,u) + \gamma \sum_{x' \in \mathcal{X}} \bar{V}(x') \bar{M}^*(x,u,x'),
    \end{aligned}
\end{equation}
with the regularized model optimization given by
{\small
\begin{equation}
    \label{eq:regularized_model_optimization_BMDP}
    \begin{aligned}
        \underaccent{\bar}{M}^*(x,u,x') & = \arg\min_{M \in \mathcal{P}} \sum_{x' \in \mathcal{X}} \underaccent{\bar}{V}(x') M(x,u,x') + \lambda d(M,M_E), \\
        \bar{M}^*(x,u,x') & = \arg\max_{M \in \mathcal{P}} \sum_{x' \in \mathcal{X}} \bar{V}(x') M(x,u,x') - \lambda d(M,M_E).
    \end{aligned}
\end{equation}
}
We propose to select $\lambda=0$ in the absence of data ($T(x,u)=0$) and otherwise as a function of system properties and the number of observed transitions from every state-action pair, such that
\begin{equation}
    \lambda(x,u) = c \sqrt{\frac{T(x,u)}{\log{\left(|\mathcal{X}||\mathcal{U}|/\delta\right)}}},
\end{equation}
where $c$ is a constant, and $\delta$ is a confidence parameter. This formulation is based on Hoeffding's inequality~\cite{Hoeffding1963}. This usage is similar to the PAC-MDP framework~\cite{Strehl2009}, as implemented, e.g., in the UCRL2 algorithm~\cite{Jaksch2010}.

The algorithm presented in this section is a straightforward extension of the tabular Q-learning algorithm, where we can use the Q-function bounds to guide the agent's exploration. The Q-bound updates are computationally efficient: without regularization, they reduce to simple min/max operations over the model set bounds, while with regularization, the optimization remains convex and tractable. The computational complexity scales similarly to other finite MDP methods like UCRL2 or PSRL, with runtime dominated by the state-action space size rather than the model set optimization. Note that all the convergence properties of Q-learning discussed in the previous sections are retained. In some cases, we can even guarantee finite-time convergence of the exploring policy to the optimal policy, as shown in the next section.

For each state-action pair, the optimization in~\eqref{eq:regularized_model_optimization_BMDP} minimizes a separable convex function over a box intersected with the probability simplex. Such singly-constrained separable problems have a dual that reduces to a scalar equation in the multiplier of the sum constraint~\cite{Patriksson2008}, so a bisection on this multiplier replaces a general conic solver. The toolbox solves all state-action pairs of a sweep as one vectorized batch, which reduces the cost of a Q-bound sweep by about two orders of magnitude.

\subsection{Finite-Time Convergence}\label{subsec:finite_time}
We can guarantee finite-time convergence of the proposed exploration strategy under the following conditions:
\begin{assumption}\label{assump:finite_time}
    For every $(x,u)\in \mathcal{X}\times\mathcal{U}$, one of the following holds:
    \begin{enumerate}
        \item \textbf{Deterministic transitions:} For all $x'\in \mathcal{X}$, $M(x,u,x')\in \{0,1\}$.
        \item \textbf{Tight bounds:} For all $x'\in \mathcal{X}$, $\underaccent{\bar}{M}(x,u,x') = \bar{M}(x,u,x')$.
    \end{enumerate}
    Furthermore, $\lambda(x,u)$ is set to rely entirely on the data for the deterministic case, that is, $\lambda(x,u)=\infty$ if 1) holds and $T(x,u)>0$.
\end{assumption}

The finite-time convergence result is formalized in the following lemma.
\begin{lemma}\label{lem:finite_time_convergence}
    Let $\pi^\phi$ be the exploring policy defined in~\eqref{eq:exploring_policy} and let $Q$ be the Q-function from {Algorithm~\ref{alg:Q-learning_regularized}}, using model set $\hat{\mathcal{M}}$ and divergence $d$ with $d(P,P_E)\geq0$ and $d(P,P_E)=0$ if and only if $P=P_E$. Assume deterministic rewards. Assume that every state is visited infinitely often (e.g., via exploring starts), and that {Assumption~\ref{assump:finite_time}} holds. Then, the Q-bounds converge to $Q^*$ in finite time with probability 1, and the $\pi^\phi$ converges to the optimal policy in finite time.
\end{lemma}
\begin{proof}
    Consider any state-action pair $(x,u)$. The exploring policy ensures that any state-action pair with uncertain Q-bounds is eventually sampled at least once (in finite time). Under {Assumption~\ref{assump:finite_time}}, we have:
    \begin{enumerate}
        \item In the \emph{deterministic case}, a finite number of visits yields the exact empirical kernel $P_E(x,u,\cdot)$; with $\lambda(x,u)=\infty$, the optimization in~\eqref{eq:regularized_model_set_optimization} forces $P=P_E$, making the Q-bounds exact in finite time.
        \item In the \emph{tight-bound case}, the Q-bound update reduces to the standard Bellman update, since the model set contains only a single model.
    \end{enumerate}
    The reward function set $\mathcal{G}$ is reduced to the true reward function $g$ in finite time, since we have finite state and action spaces. The model uncertainty is thus completely eliminated. By {Theorem~\ref{thm:exploration_and_regularization_convergence}}, the exploring policy therefore converges to the optimal policy in finite time with probability 1.
\end{proof}

Although this paper aims to present a general theoretical framework that can be cast into both finite and infinite state/action spaces, formulating a practical algorithm for the infinite-state, infinite-action case is considerably more challenging. In practice, one must resort to function approximation techniques to parameterize the Q-function. Such a parameterization typically leads to difficulties when performing optimistic max-max optimization because the overestimation inherent in the optimistic operator can result in divergence issues or unstable learning dynamics~\cite{Tsitsiklis1997}. The development of such an algorithm is left for future work.

A practical workaround is to construct a finite abstraction of the original system. By discretizing the state and action spaces, one can apply the proposed exploration strategy and Q-bound updates in a tractable manner, with approximation accuracy improving as the discretization becomes finer. This approach faces the same computational scaling challenges as other state-of-the-art finite MDP RL methods such as the ones used in the comparison in the next section.

\section{RESULTS}\label{sec:results}
We apply the proposed practical algorithm to three benchmark reinforcement learning environments: Frozen Lake, Cartpole, and Taxi from the OpenAI Gym~\cite{Brockman2016}.

\subsection{Frozen Lake Environment}\label{subsec:frozenlake}
In the Frozen Lake environment, the agent navigates a gridworld to avoid obstacles and reach a goal state, receiving a reward of 1 upon success and 0 otherwise. The transitions are stochastic due to the slippery nature of the ice, with the agent moving in either the intended direction or one of the two perpendicular directions, each with probability $1/3$.

\begin{figure}[!t]
    \centering
    \includegraphics[width=\columnwidth]{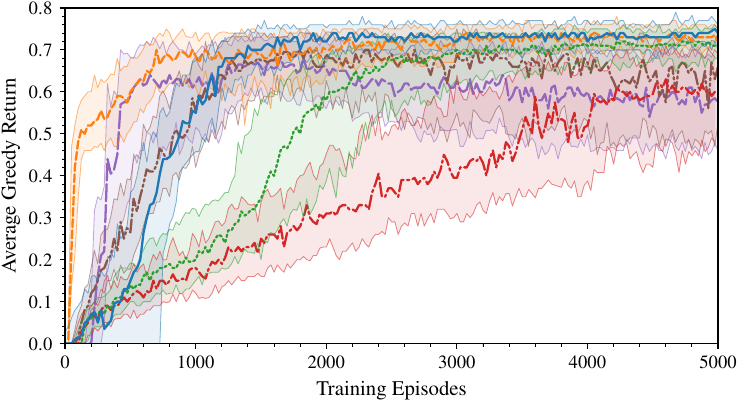}
    \caption{Evaluation performance of BUMEX with $L=\infty$ (\protect\legsym{mplblue}) and $L=100$ (\protect\legsym{mplorange}[dashed]) versus standard $\epsilon$-greedy Q-learning (\protect\legsym{mplgreen}[dotted]), Q-learning with UCB exploration (\protect\legsym{mplred}[dash dot]), UCRL2 (\protect\legsym{mplpurple}[loosely dashed]), and PSRL (\protect\legsym{mplbrown}[dash dot dot]) in the Frozen Lake environment. Plotted are the median, and the 25th and 75th percentiles of 100 Monte Carlo runs, where the performance is evaluated every 25 training episodes by averaging the returns obtained by applying the greedy policy for 100 episodes.}
    \label{fig:FrozenLake}
\end{figure}

We define the model set to include knowledge about adjacency in the gridworld. In particular, we assume that all transitions for which the true model has probability zero are known, while all others have probabilities between 0 and 1. The reward function set $\mathcal{G}$ contains only the true reward function $g$, assuming full knowledge of the rewards. This model set definition is similar to what a human player might assume when playing the game, i.e., the agent knows the layout of the gridworld and the locations of obstacles and goals, but not the exact transition probabilities. The results are shown in Fig.~\ref{fig:FrozenLake}.

We compare standard (randomly exploring) $\epsilon$-greedy Q-learning, Q-learning with UCB1 exploration~\cite{Jin2018}, UCRL2~\cite{Jaksch2010}, and PSRL~\cite{Osband2013} to two variants of BUMEX: one without regularization ($L=\infty$) and one with regularization ($L=100$). All experiments use $\xi=1$, $\zeta=0$, $\beta$ as in~\eqref{eq:EI}, $c=5$, $\delta=0.05$, $\alpha=0.05$, and $\gamma=0.95$, with $\epsilon$ decaying exponentially from 1 to 0.01. The results indicate that BUMEX converges to the optimal policy quickly and consistently, even when the Q-bound iterations are only performed at the beginning. When performing Q-bound iterations (with regularization) every 100 episodes, BUMEX converges even faster, although this comes at an additional computational cost induced by the repeated calculation of Bellman operators. Note that methods such as UCRL2 and PSRL also rely on repeated Bellman operations with comparable computational costs.

\subsection{Cartpole Environment}
In the Cartpole environment, the agent is tasked with balancing a pole on a cart by moving the cart left or right, receiving a reward of 1 per time step the pole is balanced. The transitions are deterministic and governed by the physics of the cartpole system. We create finite state and action spaces by discretizing the continuous state and action spaces, which introduces stochasticity into the transition model. Furthermore, we perform quadratic reward shaping during the training phase to speed up learning.

\begin{figure}[!t]
    \centering
    \includegraphics[width=\columnwidth]{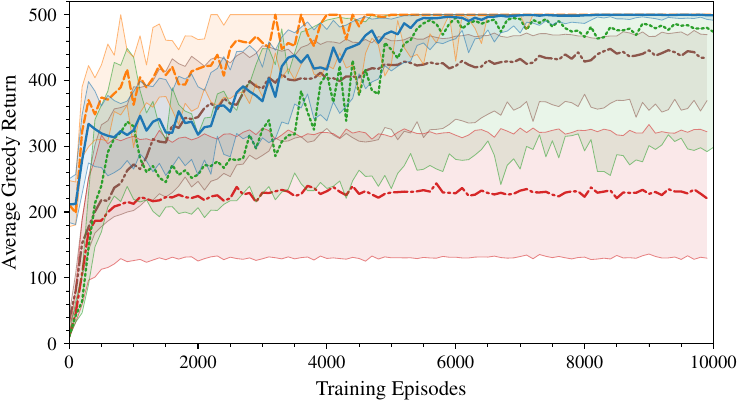}
    \caption{Evaluation performance of BUMEX with $L=\infty$ (\protect\legsym{mplblue}) and $L=200$ (\protect\legsym{mplorange}[dashed]) versus standard $\epsilon$-greedy Q-learning (\protect\legsym{mplgreen}[dotted]), Q-learning with UCB exploration (\protect\legsym{mplred}[dash dot]), and PSRL (\protect\legsym{mplbrown}[dash dot dot]) in the Cartpole environment. Plotted are the median, and the 25th and 75th percentiles of 100 Monte Carlo runs, where the performance is evaluated every 100 training episodes by averaging the returns obtained by applying the greedy policy for 100 episodes.}
    \label{fig:Cartpole}
\end{figure}

We define the model set to include knowledge about the cartpole dynamics, i.e., the cartpole is governed by Newtonian physics. In particular, we assume that the dynamics equations are known up to an unknown mass parameter. We obtain the set $\mathcal{P}$ by looping over the set of possible mass values, obtaining the transition probability tensor for each, then taking the max and min over these tensors to obtain $\underaccent{\bar}{M}$ and $\bar{M}$ in~\eqref{eq:BMDP_probability_transitions}. The reward function set $\mathcal{G}$ contains only the true reward function $g$, i.e., we assume full knowledge of the rewards. The results are shown in Fig.~\ref{fig:Cartpole}.

We compare the algorithms $\epsilon$-greedy, UCB1, PSRL, BUMEX with $L=\infty$ and $L=200$, with the same $\xi, \zeta, \beta$, $c$, and $\delta$ as in Section~\ref{subsec:frozenlake} (UCRL2 is excluded due to poor performance). We use $\alpha=0.03$, $\gamma=0.97$, and $\epsilon$ decaying exponentially from 1 to 0.001. The results are similar to those in the Frozen Lake environment, showing that the regularized method converges fastest and most consistently, followed by the non-regularized version. The more limited performance gain from regularization in this environment can be attributed to the larger state space dimension.

\subsection{Taxi Environment}
The Taxi environment consists of a more complicated gridworld, in which the agent is tasked with picking up and dropping off passengers at dedicated locations. The agent is rewarded -1 at every timestep, -10 for illegal passenger pickups/dropoffs, and +20 for a successful passenger dropoff. The gridworld contains walls which the agent cannot move through. Note that this environment is deterministic.

Our prior model knowledge largely takes the same form as in Section~\ref{subsec:frozenlake}, i.e., we assume that the agent knows the adjacency structure of the gridworld and the locations of the pickup and dropoff points. We assume the agent does not have knowledge of wall placements or the reward function structure.

\begin{figure}[!t]
    \centering
    \includegraphics[width=\columnwidth]{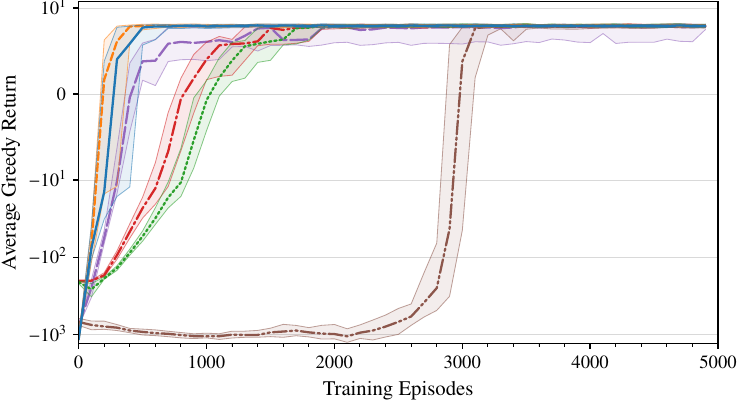}
    \caption{Evaluation performance of BUMEX with $L=\infty$ (\protect\legsym{mplblue}) and $L=100$ (\protect\legsym{mplorange}[dashed]) versus standard $\epsilon$-greedy Q-learning (\protect\legsym{mplgreen}[dotted]), Q-learning with UCB exploration (\protect\legsym{mplred}[dash dot]), UCRL2 (\protect\legsym{mplpurple}[loosely dashed]), and PSRL (\protect\legsym{mplbrown}[dash dot dot]) in the Taxi environment. Plotted are the median, and the 25th and 75th percentiles of 100 Monte Carlo runs, where the performance is evaluated every 100 training episodes by averaging the returns obtained by applying the greedy policy for 100 episodes.}
    \label{fig:Taxi}
\end{figure}

We compare the same algorithms ($\epsilon$-greedy, UCB1, UCRL2, PSRL, BUMEX with $L=\infty$ and $L=100$) as in the previous sections, with the same $\xi, \zeta, \beta$, $c$, and $\delta$. We use $\alpha=0.03$, $\gamma=0.98$, and $\epsilon$ decaying exponentially from 1 to 0.01. The results are shown in Fig.~\ref{fig:Taxi} and are consistent with the other two environments. This experiment specifically highlights the benefits of having accurate dynamics knowledge while learning unknown rewards. The proposed method converges very quickly once the deterministic reward function has been learned.

\section{CONCLUSIONS AND FUTURE WORK}\label{sec:conclusions}
In this paper, we proposed an exploration strategy for reinforcement learning that leverages model-based Q-function bounds to guide the agent's exploration. Our work establishes multiple theoretical results that guarantee the convergence of the proposed exploration strategy to the optimal policy in a general MDP setting. Furthermore, in the finite state and action space case, and under reasonable assumptions on the model set, we obtain a practical algorithm, BUMEX, with the same convergence guarantees. We also demonstrated that, under additional assumptions on the transition probabilities, our method achieves finite-time convergence to the optimal policy. The effectiveness of the proposed strategy was validated in several Gym environments, where it outperformed standard exploration methods.

Future work includes extending the theoretical results to infinite state-action spaces, e.g., through function approximation. Additionally, a sample complexity analysis of BUMEX is desired to quantify the efficiency compared to state-of-the-art RL methods.

\bibliographystyle{IEEEtran.bst}
{\small
\bibliography{library.bib}
}

\end{document}